\documentclass{llncs}
\usepackage{graphicx}
\usepackage{multirow}
\usepackage{todonotes}
\usepackage{stmaryrd}
\usepackage{amsmath}
\usepackage{amssymb}
\usepackage{hyperref}
\usepackage{subfig}

\renewcommand{\vec}[1]{\boldsymbol{#1}}
\newcommand{\given}{\, | \,}
\newcommand{\hath}{\widehat{h}}

\newcommand{\haty}{\widehat{y}}

\newcommand{\fromto}{\longrightarrow}

\newcommand*{\defeq}{\mathrel{\vcenter{\baselineskip0.5ex \lineskiplimit0pt
			\hbox{\footnotesize.}\hbox{\footnotesize.}}}%
	=}
	
\newcommand{\svert}{\, \vert \, }
\newcommand{\cX}{\mathcal{X}}
\newcommand{\cY}{\mathcal{Y}}
\newcommand{\cH}{\mathcal{H}}
\newcommand{\cD}{\mathcal{D}}
\newcommand{\Prob}{P}
\newcommand{\prob}{p}
\newcommand{\argmin}{\operatorname*{argmin}}

\begin{document}
\title{\LARGE Aleatoric and Epistemic Uncertainty\\ with Random Forests}
\titlerunning{Aleatoric and Epistemic Uncertainty with Random Forests}
%
\author{Mohammad Hossein Shaker\inst{1} \and
Eyke H\"{u}llermeier\inst{1}}
%

\institute{Heinz Nixdorf Institute and Department of Computer Science\\
 Paderborn University, Germany\\
\email{\{mhshaker,eyke\}@upb.de}
}
\maketitle              
\begin{abstract}
Due to the steadily increasing relevance of machine learning for practical applications, many of which are coming with safety requirements, the notion of uncertainty has received increasing attention in machine learning research in the last couple of years. In particular, the idea of distinguishing between two important types of uncertainty, often refereed to as \emph{aleatoric} and \emph{epistemic}, has recently been studied in the setting of supervised learning. In this paper, we propose to quantify these uncertainties with random forests. More specifically, we show how two general approaches for measuring the learner's aleatoric and epistemic uncertainty in a prediction can be instantiated with decision trees and random forests as learning algorithms in a classification setting. In this regard, we also compare random forests with deep neural networks, which have been used for a similar purpose.  
\keywords{machine learning \and uncertainty \and random forest}
\end{abstract}

\section{Introduction}

The notion of uncertainty has received increasing attention in machine learning research in the last couple of years, especially due to the steadily increasing relevance of machine learning for practical applications. In fact, a trustworthy representation of uncertainty should be considered as a key feature of any machine learning method, all the more in safety-critical application domains such as medicine \cite{yang_ur09,lamb_rc11} or socio-technical systems \cite{vars_es16,vars_ot16}. 

In the general literature on uncertainty, a distinction is made between two inherently different sources of uncertainty, which are often referred to as \emph{aleatoric} and \emph{epistemic} \cite{hora_aa96}. Roughly speaking, aleatoric (\emph{aka} statistical) uncertainty refers to the notion of randomness, that is, the variability in the outcome of an experiment which is due to inherently random effects. The prototypical example of aleatoric uncertainty is coin flipping. 
As opposed to this, epistemic (\emph{aka} systematic) uncertainty refers to uncertainty caused by a lack of knowledge, i.e., it relates to the epistemic state of an agent or decision maker. This uncertainty can in principle be reduced on the basis of additional information. In other words, epistemic uncertainty refers to the \emph{reducible} part of the (total) uncertainty, whereas aleatoric uncertainty refers to the \emph{non-reducible} part. 

More recently, this distinction has also received attention in machine learning, where the ``agent'' is a learning algorithm \cite{mpub272}. In particular, a distinction between aleatoric and epistemic uncertainty has been advocated in the literature on deep learning \cite{kend_wu17}, where the limited awareness of neural networks of their own competence has been demonstrated quite nicely. For example, experiments on image classification have shown that a trained model does often fail on specific instances, despite being very confident in its prediction. Moreover, such models are often lacking robustness and can easily be fooled by ``adversarial examples'' \cite{pape_dk18}: Drastic changes of a prediction may already be provoked by minor, actually unimportant changes of an object. This problem has not only been observed for images but also for other types of data, such as natural language text \cite{sato_ia18}.  

In this paper, we advocate the use of decision trees and random forests, not only as a powerful machine learning method with state-of-the-art predictive performance, but also for measuring and quantifying predictive uncertainty. More specifically, we show how two general approaches for measuring the learner's aleatoric and epistemic uncertainty in a prediction (recalled in Section 2) can be instantiated with decision trees and random forests as learning algorithms in a classification setting (Section 3). In an experimental study on uncertainty-based abstention (Section 4), we compare random forests with deep neural networks, which have been used for a similar purpose.

\section{Epistemic and Aleatoric Uncertainty}\label{sec:EpisAlea}

We consider a standard setting of supervised learning, in which a learner is given access to a set of (i.i.d.) training data $\mathcal{D} \defeq \{ (\vec{x}_i , y_i )\}_{i=1}^N \subset \mathcal{X} \times \mathcal{Y}$, where $\mathcal{X}$ is an instance space and $\mathcal{Y}$ the set of outcomes that can be associated with an instance. In particular, we focus on the classification scenario, where $\cY=\{y_1, \ldots, y_K\}$ consists of a finite set of class labels, with binary classification ($\cY = \{0,1\}$) as an important special case.

Suppose a \emph{hypothesis space} $\mathcal{H}$ to be given, where a hypothesis $h \in \cH$ is a mapping $\cX \fromto \mathbb{P}(\cY)$, i.e., a hypothesis maps instances $\vec{x}\in\cX$ to probability distributions on outcomes. The goal of the learner is to induce a hypothesis $h^* \in \mathcal{H}$ with low risk (expected loss)
\begin{equation}
R(h) \defeq \int_{\cX \times \cY} \ell( h(\vec{x}) , y) \, d \, \Prob(\vec{x} , y) \enspace ,
\end{equation}
where $P$ is the (unknown) data-generating process (a probability distribution on $\cX \times \cY$), and $\ell: \, \mathcal{Y} \times \mathcal{Y} \longrightarrow \mathbb{R}$ a loss function. 
This choice of a hypothesis is commonly guided by the empirical risk 
\begin{equation}
R_{emp}(h) \defeq  \frac{1}{N} \sum_{i=1}^N \ell(h(\vec{x}) , y) \enspace ,
\end{equation}
i.e., the performance of a hypothesis on the training data. However, since $R_{emp}(h)$ is only an estimation of the true risk $R(h)$, the empirical risk minimizer (or any other predictor)
\begin{equation}\label{eq:argerm}
\hath \defeq \argmin_{h \in \cH} R_{emp}(h)
\end{equation}
favored by the learner will normally not coincide with the true risk minimizer (Bayes predictor)
\begin{equation}\label{eq:bayespred}
h^* \defeq \argmin_{h \in \cH} R(h) \, .
\end{equation} 
Correspondingly, there remains uncertainty regarding $h^*$ as well as the approximation quality of $\hath$ (in the sense of its proximity to $h^*$) and its true risk $R(\hath)$.

Eventually, one is often interested in the \emph{predictive uncertainty}, i.e., the uncertainty related to the prediction $\haty_{q}$ for a concrete query instance $\vec{x}_{q} \in \cX$. In other words, given a partial observation $(\vec{x}_{q} , \cdot)$, we are wondering what can be said about the missing outcome, especially about the uncertainty related to a prediction of that outcome. Indeed, estimating and quantifying uncertainty in a transductive way, in the sense of tailoring it to individual instances, is arguably important and practically more relevant than a kind of average accuracy or confidence, which is often reported in machine learning. 



\begin{figure}
\begin{center}
\includegraphics[scale=0.35]{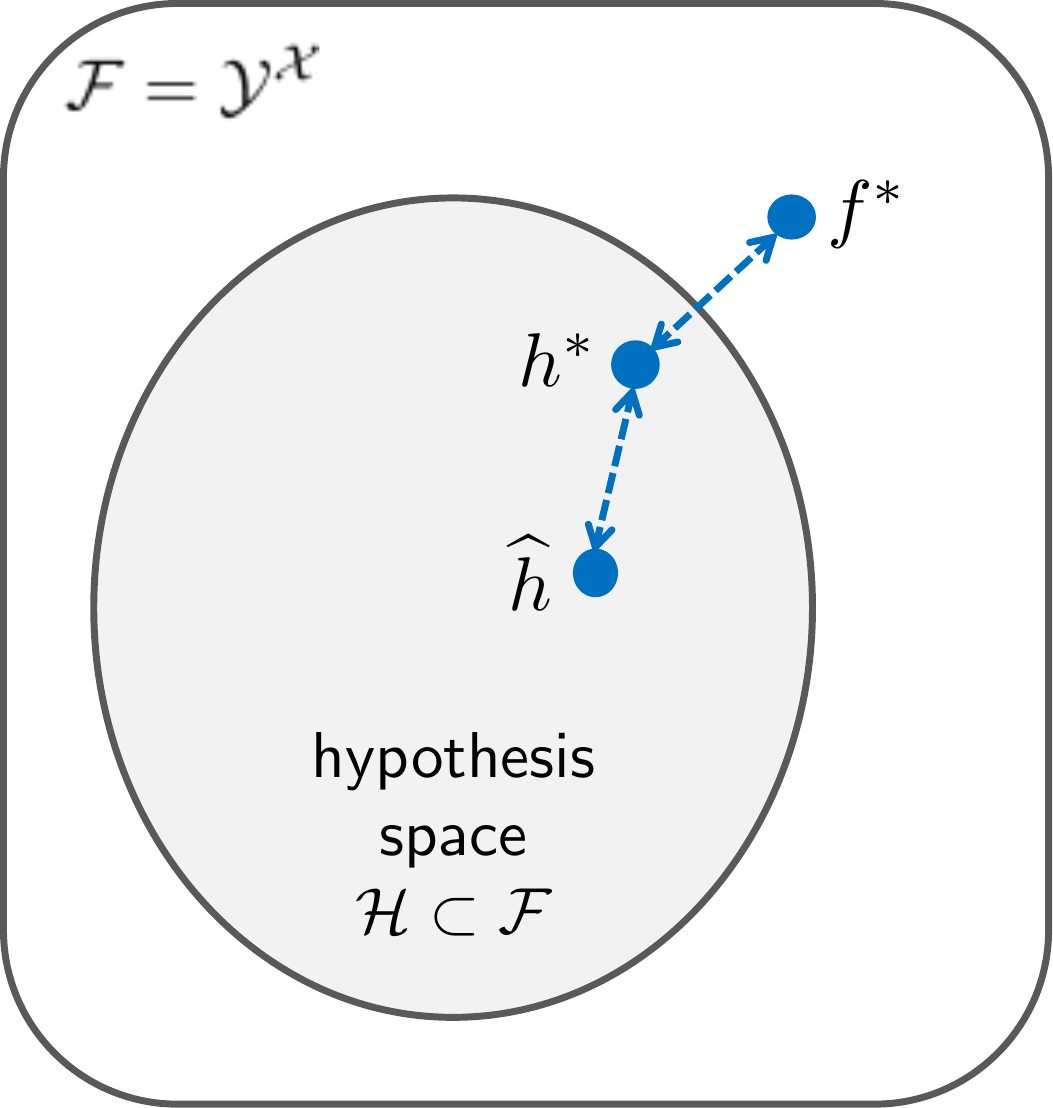} \qquad
\caption{Different types of uncertainties related to different types of discrepancies and approximation errors:  $f^*$ is the pointwise Bayes predictor, $h^*$ is the best predictor within the hypothesis space, and $\hath$ the predictor produced by the learning algorithm.}
\label{fig:approx}
\end{center}
\end{figure}

As the prediction $\haty_{q}$ constitutes the end of a process that consists of different learning and approximation steps, all errors and uncertainties related to these steps may also contribute to the uncertainty about $\haty_{q}$ (cf.\ Fig.\ \ref{fig:approx}):
\begin{itemize}
\item Since the dependency between $\cX$ and $\cY$ is typically non-deterministic, the description of a new prediction problem in the form of an instance $\vec{x}_{q}$ gives rise to a conditional probability distribution
\begin{equation}\label{eq:ccp}
\prob( y \given \vec{x}_{q}) = \frac{\prob(\vec{x}_{q} , y)}{\prob(\vec{x}_q)} 
\end{equation}
on $\cY$, but it does normally not identify a single outcome $y$ in a unique way. Thus, even given full information in the form of the measure $\Prob$ (and its density $\prob$), uncertainty about the actual outcome $y$ remains. This uncertainty is of an \emph{aleatoric} nature. In some cases, the distribution (\ref{eq:ccp}) itself (called the predictive posterior distribution in Bayesian inference) might be delivered as a prediction. Yet, when having to commit to a point estimate, the best prediction (in the sense of minimizing the expected loss) is prescribed by the pointwise Bayes predictor $f^*$, which is defined by
\begin{equation}\label{eq:pointbayespred}
f^*(\vec{x}) \defeq \argmin_{\haty \in \cY} \int_\cY \ell(y , \haty) \, d \Prob( y \given \vec{x} )
\end{equation} 
for each $\vec{x} \in \cX$.

\item The Bayes predictor (\ref{eq:bayespred}) does not necessarily coincide with the pointwise Bayes predictor (\ref{eq:pointbayespred}). This discrepancy between $h^*$ and $f^*$ is connected to the uncertainty regarding the right type of model to be fit, and hence the choice of the hypothesis space $\cH$. We refer to this uncertainty as \emph{model uncertainty}. Thus, due to this uncertainty, one can not guarantee that $h^*(\vec{x}) = f^*(\vec{x})$, or, in case the hypothesis $h^*$ delivers probabilistic predictions $\prob(y \given h^*, \vec{x})$ instead of point predictions, that $\prob( \cdot \given h^*, \vec{x}) = \prob( \cdot \given \vec{x})$.

\item The hypothesis $\hath$ produced by the learning algorithm, for example the empirical risk minimizer (\ref{eq:argerm}), is only an estimate of $h^*$, and the quality of this estimate strongly depends on the quality and the amount of training data. We refer to the discrepancy between $\hath$ and $h^*$, i.e., the uncertainty about how well the former approximates the latter, as \emph{approximation uncertainty}. 

\end{itemize}
As already said, aleatoric uncertainty is typically understood as uncertainty that is due to influences on the data-generating process that are inherently random, that is, due to the non-deterministic nature of the sought input/output dependency. This part of the uncertainty is irreducible, in the sense that the learner cannot get rid of it. Model uncertainty and approximation uncertainty, on the other hand, are subsumed under the notion of epistemic uncertainty, that is, uncertainty due to a lack of knowledge about the perfect predictor (\ref{eq:pointbayespred}). Obviously, this lack of knowledge will strongly depend on the underlying hypothesis space $\cH$ as well as the amount of data seen so far: The larger the number $N = | \mathcal{D}|$ of observations, the less ignorant the learner will be when having to make a new prediction. In the limit, when $N \rightarrow \infty$, a consistent learner will be able to identify $h^*$. Moreover, the ``larger'' the hypothesis pace $\cH$, i.e., the weaker the prior knowledge about the sought dependency, the higher the epistemic uncertainty will be, and the more data will be needed to resolve this uncertainty.


How to capture these intuitive notions of aleatoric and epistemic uncertainty in terms of quantitative measures? In the following, we briefly recall two proposals that have recently been made in the literature.  

\subsection{Entropy Measures}
\label{sec:entropy}

An attempt at measuring and separating aleatoric and epistemic uncertainty on the basis of classical information-theoretic measures of entropy is made in \cite{depe_du18}. This approach is developed in the context of neural networks for regression, but the idea as such is more general and can also be applied to other settings. A similar approach was recently adopted in \cite{mobi_dc19}. 

More specifically, given a query instance $\vec{x}$, the idea is to measure the total uncertainty in a prediction in terms of the (Shannon) entropy of the predictive posterior distribution, which, in the case of discrete $\cY$, is given as 
\begin{equation}\label{eq:eto}
H \big[ \, \prob(y \given \vec{x}) \, \big] = \mathbf{E}_{\prob(y \given \vec{x})} \big\{ - \log_2 \prob(y \given \vec{x})  \big\}  = - \sum_{y \in \cY} \prob(y \given \vec{x}) \log_2 \prob(y \given \vec{x}) \, .
\end{equation}
Moreover, the epistemic uncertainty is measured in terms of the mutual information between hypotheses and outcomes (i.e., the Kullback-Leibler divergence between the joint distribution of outcomes and hypotheses and the product of their marginals):
\begin{equation}\label{eq:eep}
I(y, h)  = \mathbf{E}_{p(y,h)} \left\{  \log_2 \left( \frac{p(y,h)}{p(y) p(h)} \right) \right\} \, ,
\end{equation}
Finally, the aleatoric uncertainty is specified in terms of the difference between (\ref{eq:eto}) and (\ref{eq:eep}), which is given by
\begin{equation}\label{eq:eal}
\mathbf{E}_{p( h \given \mathcal{D})} H \big[ \prob(y \given h, \vec{x}) \big] = 
- \int_{\cH} p( h \given \mathcal{D}) \left( \sum_{y \in \cY} \prob(y \given h, \vec{x}) \log_2 \prob(y \given h, \vec{x}) \right) \, d \, h
\end{equation}
The idea underlying (\ref{eq:eal}) is as follows: By fixing a hypothesis $h \in \cH$, the epistemic uncertainty is essentially removed. Thus, the entropy $H [ \prob(y \given h, \vec{x})]$, i.e., the entropy of the conditional distribution on $\cY$ predicted by $h$ for the query instance $\vec{x}$, is a natural measure of the aleatoric uncertainty. However, since $h$ is not precisely known, aleatoric uncertainty is measured in terms of the expectation of this entropy with regard to the posterior probability $p( h \given \mathcal{D})$. 

The epistemic uncertainty (\ref{eq:eep}) captures the dependency between the probability distribution on $\cY$ and the hypothesis $h$. Roughly speaking, (\ref{eq:eep}) is high if the distribution $\prob(y \given h, \vec{x})$ varies a lot for different hypotheses $h$ with high probability. This is plausible, because the existence of different hypotheses, all considered (more or less) probable but leading to quite different predictions, can indeed be seen as a sign for high epistemic uncertainty. 

Obviously, (\ref{eq:eep}) and (\ref{eq:eal}) cannot be computed efficiently, because they involve an integration over the hypothesis space $\cH$. One idea, therefore, is to approximate these measures by means of ensemble techniques \cite{mobi_dc19}, that is, to represent the posterior distribution $p( h \given \mathcal{D})$ by a finite ensemble of hypotheses $H = \{ h_1, \ldots , h_M \}$. An approximation of (\ref{eq:eal}) can then be obtained by
\begin{equation}\label{eq:approxua}
u_a(\vec{x}) \defeq
- \frac{1}{M} \sum_{i=1}^M 
\sum_{y \in \cY} \prob(y \given h_i, \vec{x}) \log_2 \prob(y \given h_i, \vec{x})   \, ,
\end{equation}
an approximation of (\ref{eq:eto}) by
\begin{equation}\label{eq:approxut}
u_t(\vec{x}) \defeq  - \sum_{y \in \cY} \left( \frac{1}{M} \sum_{i=1}^M \prob(y \given h_i, \vec{x}) \right) \log_2 \left( \frac{1}{M} \sum_{i=1}^M \prob(y \given h_i, \vec{x}) \right) \, ,
\end{equation} 
and finally and approximation of (\ref{eq:eep}) by $u_e(\vec{x}) \defeq u_t(\vec{x}) - u_a(\vec{x})$.
For neural networks, it has been shown that techniques such as Dropout \cite{gal_bc16} and DropConnect \cite{mobi_dc19} can be interpreted as (implicit) ensemble methods, and can hence be used to implement this approach.

\subsection{Measures based on Relative Likelihood}
\label{sec:rl}

Another approach, put forward in \cite{mpub272}, is based on the use of relative likelihoods, historically proposed by \cite{birnbaum:1962} and then justified in other settings such as possibility theory~\cite{walley:1999}. Here, we briefly recall this approach for the case of binary classification, i.e., where $\cY = \{ 0,1 \}$; see \cite{nguyen:2018} for an extension to the case of multinomial classification.

Given training data $\cD = \{ (\vec{x}_i , y_i) \}_{i=1}^N \subset \cX \times \cY$, the normalized likelihood of $h \in \cH$ is defined as
\begin{align}\label{eq:plausibility}
\pi_{\mathcal{H}}(h) \defeq \frac{L(h)}{L(h^{ml})} =  
\frac{L(h)}{\max_{h' \in \mathcal{H}} L(h')} \enspace ,
\end{align}  
where $L(h) = \prod_{i=1}^N \prob(y_i \svert h, \vec{x}_i)$ is the likelihood of $h$, and $h^{ml} \in \mathcal{H}$ the maximum likelihood estimation.
For a given instance $\vec{x}$, the degrees of support (plausibility) of the two classes are defined as follows: 
\begin{eqnarray}
\pi(1\svert \vec{x}) & =  & \sup_{h \in \mathcal{H}} \min \big[\pi_{\mathcal{H}}(h), \prob(1 \svert h, \vec{x}) - \prob(0 \svert h, \vec{x}) \big], \label{eq:support posx}\\
\pi(0 \svert \vec{x}) & = & \sup_{h \in \mathcal{H}} \min \big[\pi_{\mathcal{H}}(h), \prob(0 \svert h, \vec{x}) - \prob(1 \svert h, \vec{x}) \big]. \label{eq:support negx}
\end{eqnarray} 
So, $\pi(1 \svert  \vec{x})$ is high if and only if a highly plausible hypothesis supports the positive class much stronger (in terms of the assigned probability) than the negative class (and $\pi(0 \svert \vec{x})$ can be interpreted analogously).
Given the above degrees of support, the degrees of epistemic and aleatoric uncertainty are defined as follows:
\begin{eqnarray}
u_e(\vec{x}) & = & \min \big[ \pi(1  \svert \vec{x}), \pi(0 \svert \vec{x}) \big] \, , \label{eq:epistemic} \\
u_a(\vec{x}) & = & 1 - \max \big[ \pi(1  \svert \vec{x}), \pi(0  \svert \vec{x}) \big] \, .\label{eq:aleatoric}
\end{eqnarray}  
Thus, epistemic uncertainty refers to the case where both the positive and the negative class appear to be plausible, while the degree of aleatoric uncertainty \eqref{eq:aleatoric} is the degree to which none of the classes is supported. More specifically, the above measures have the following properties: 
\begin{itemize}
\item[-] $u_e(\vec{x})$ will be high if class probabilities strongly vary within the set of plausible hypotheses, i.e., if we are unsure how to compare these probabilities. In particular, it will be $1$ if and only if we have $h(\vec{x})=1$ and $h'(\vec{x})=0$ for two totally plausible hypotheses $h$ and $h'$; 
\item[-] $u_a(\vec{x})$ will be high if class probabilities are similar for all plausible hypotheses, i.e., if there is strong evidence that $h(\vec{x}) \approx 0.5$. In particular, it will be close to $1$ if all plausible hypotheses allocate their probability mass around $h(\vec{x})=0.5$.
\end{itemize}    
As can be seen, the measures (\ref{eq:epistemic}) and (\ref{eq:aleatoric}) are actually quite similar in spirit to the measures (\ref{eq:eep}) and (\ref{eq:eal}).

\section{Random Forests}
\label{sec:rf}

Our basic idea is to instantiate the (generic) uncertainty measures presented in the previous section by means of decision trees \cite{quinlan:1986,safavian:1991}, that is, with decision trees as an underlying hypothesis space $\cH$. This idea is motivated by the fact that, firstly, decision trees can naturally be seen as probabilistic predictors \cite{krup_pe14}, and secondly, they can easily be used as an ensemble in the form of a random forest\,---\,recall that ensembling is needed for the (approximate) computation of the entropy-based measures in Section \ref{sec:entropy}.

\subsection{Entropy Measures}

The approach in Section \ref{sec:entropy} can be realized with decision forests in a quite straightforward way. Let $H= \{ h_1, \ldots, h_M \}$ be a classifier ensemble in the form of a random forest consisting of decision trees $h_i$. Moreover, recall that a decision tree $h_i$ partitions the instance space $\cX$ into (rectangular) regions $R_{i,1}, \ldots , R_{i,L_i}$ (i.e., $\bigcup_{l=1}^{L_i} R_{i,l} = \mathcal{X}$ and $R_{i,k} \cap R_{i,l} = \emptyset$ for $k \neq l$) associated with corresponding leafs of the tree (each leaf node defines a region $R$). Given a query instance $\vec{x}$, the probabilistic prediction produced by the tree $h_i$ is specified by the Laplace-corrected relative frequencies of the classes $y \in \cY$ in the region $R_{i,j} \ni \vec{x}$: 
$$
\prob(y \given h_i , \vec{x}) = \frac{n_{i,j}(y)+1}{n_{i,j}+ |\cY|} \, ,
$$
where $n_{i,j}$ is the number of training instances in the leaf node $R_{i,j}$, and $n_{i,j}(y)$ the number of instances with class $y$. With probabilities estimated in this way, the uncertainty degrees (\ref{eq:approxua}) and (\ref{eq:approxut}) can directly be derived. 

\subsection{Measures based on Relative Likelihood}

Instantiating the approach in Section \ref{sec:rl} essentially means computing the degrees of support (\ref{eq:support posx}--\ref{eq:support negx}), from which everything else can easily be derived. 

As already said, a decision tree partitions the instance space into several regions, each of which can be associated with a constant predictor. More specifically, in the case of binary classification, the predictor is of the form $h_\theta$, $\theta \in \Theta = [0,1]$, where $h_\theta(\vec{x}) \equiv \theta$ is the (predicted) probability $\prob(1 \given \vec{x} \in R)$ of the positive class in the region. If we restrict inference to a local region, the underlying hypothesis space is hence given by $\mathcal{H} = \{ h_\theta \given 0 \leq \theta \leq 1 \}$.

With $n$ and $p$ the number of positive and negative instances, respectively, within a region $R$, the likelihood and the maximum likelihood estimate of $\theta$ are respectively given by 
\begin{align}
L(\theta)=
\left( \begin{array}{c}
n+p \\
n \\
\end{array} \right) 
\theta^n (1-\theta)^p  \, 
\text{ and }
\theta^{ml} =\frac{n}{n+p} \, . 
\end{align}
Therefore, the degrees of support for the positive and negative classes are 
\begin{align}
\pi(1\given \vec{x}) = \sup_{\theta \in [0,1]} &\min \left( \frac{\theta^p(1 - \theta)^n}{\big(\frac{p}{n+p}\big)^p \big(\frac{n}{n+p}\big)^n} , \, 2\theta-1 \right)  \, , \label{eq:knn sup pos}\\[2mm]
\pi(0\given \vec{x}) = \sup_{\theta \in [0,1]} &\min \left( \frac{\theta^p(1 - \theta)^n}{\big(\frac{p}{n+p}\big)^p \big(\frac{n}{n+p}\big)^n}, \, 1-2\theta \right)  \, . \label{eq:knn sup neg} 
\end{align}
Solving \eqref{eq:knn sup pos} and \eqref{eq:knn sup neg} comes down to maximizing a scalar function over a bounded domain, for which standard solvers can be used. 
From (\ref{eq:knn sup pos}--\ref{eq:knn sup neg}), the epistemic and aleatoric uncertainty associated with the region $R$ can be derived according to (\ref{eq:epistemic}) and (\ref{eq:aleatoric}), respectively. For different combinations of $n$ and $p$, these uncertainty degrees can be pre-computed.
  
Note that, for this approach, the uncertainty degrees (\ref{eq:epistemic}) and (\ref{eq:aleatoric}) can be obtained for a single tree. To leverage the ensemble $H$, we average both uncertainties over all trees in the random forest. 

\section{Experiments}

The empirical evaluation of methods for quantifying uncertainty is a non-trivial problem. In fact, unlike for the prediction of a target variable, the data does normally not contain information about any sort of ``ground truth'' uncertainty. What is often done, therefore, is to evaluate predicted uncertainties \emph{indirectly}, that is, by assessing their usefulness for improved prediction and decision making. Adopting an approach of that kind, we produced \emph{accuracy-rejection curves}, which depict the accuracy of a predictor as a function of the percentage of rejections \cite{mpub170}: A classifier, which is allowed to abstain on a certain percentage $p$ of predictions, will predict on those $(1-p)$\,\% on which it feels most certain. Being able to quantify its own uncertainty well, it should improve its accuracy with increasing $p$, hence the accuracy-rejection curve should be monotone increasing (unlike a flat curve obtained for random abstention).

\subsection{Implementation Details}

For this work, we used the Random Forest Classifier from SKlearn. The number of trees within the forest is set to 50, with the maximum level of tree grows set to 10. We use bootstrapping to create diversity between the trees of the forest. 

As a baseline to compare with, we used the DropConnect model for deep neural networks as introduced in \cite{mobi_dc19}. The idea of DropConnect is similar to Dropout, but here, instead of randomly deleting neurons, we randomly delete the connections between neurons. In this model, the act of dropping the connections is also active in the test phase. In this way, the data passes through a different network on each iteration, and therefore we can compute Monte Carlo samples for each query instance. The DropConnect model is a feed forward neural network consisting of two DropConnect layers with 32 neurons and a final softmax layer for the output. The model is trained for 20 epochs with mini batch size of 32. After the training is done, we take 50 Monte Carlo samples to create an ensemble, from which the uncertainty values can be calculated.

\subsection{Results}

Due to space limitations, we show results in the form of accuracy-rejection curves for only two exemplary data sets from the UCI repository\footnote{\url{https://archive.ics.uci.edu/ml/datasets/}}, spect and diabetes\,---\,yet, very similar results were obtained for other data sets. The data is randomly split into 70\% for training and 30\% for testing, and accuracy-rejection curves are computed on the latter (the curves shown are averages over 100 repetitions). In the following, we abbreviate the aleatoric and epistemic uncertainty degrees produced by the entropy-based approach (Section \ref{sec:entropy}) and the approach based on relative likelihood (Section \ref{sec:rl}) by AU-ent, EU-ent, AU-rl, and EU-rl, respectively. 


\begin{figure}[h]
\begin{center}

\vspace*{-8mm}

  \subfloat[spect]{\includegraphics[scale=0.3]{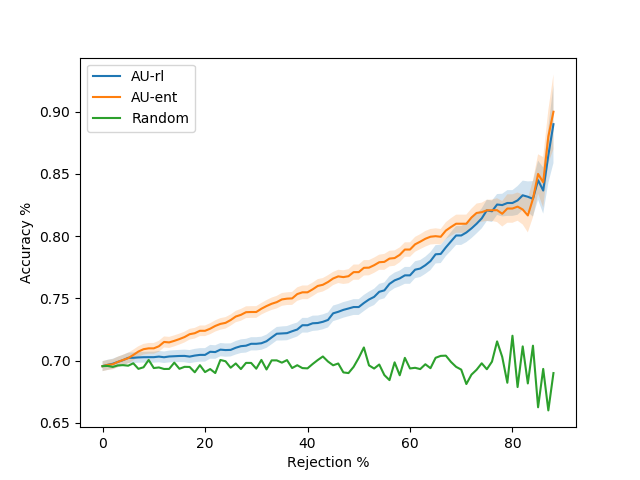} }
  \subfloat[diabetes]{\includegraphics[scale=0.3]{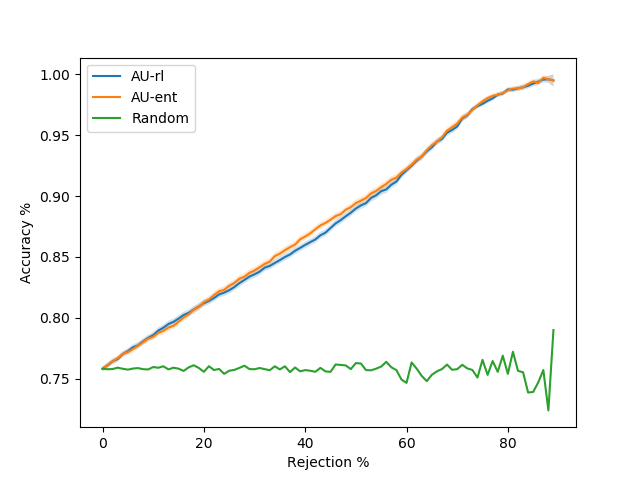}}
  
    \vspace*{-5mm}
    
    \subfloat[spect]{\includegraphics[scale=0.3]{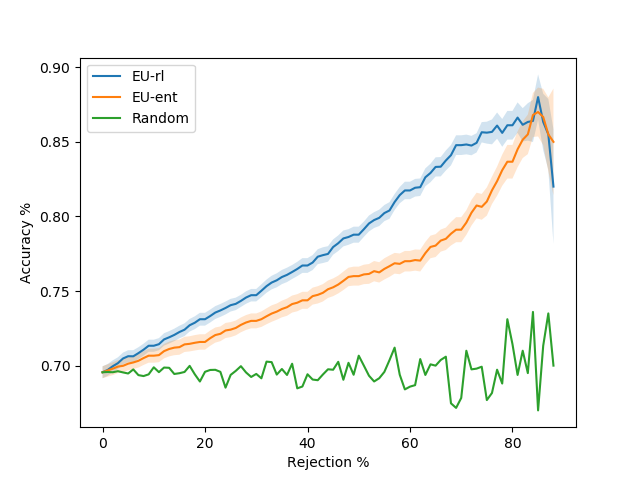} }
  \subfloat[diabetes]{\includegraphics[scale=0.3]{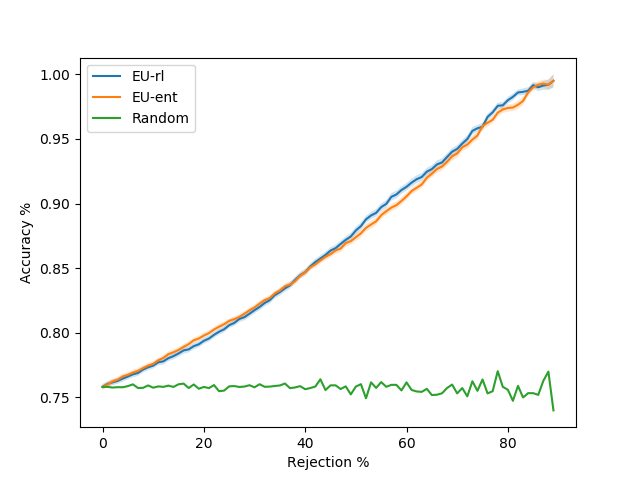}}
\caption{Accuracy-rejection curves for aleatoric (above) and epistemic (below) uncertainty using random forests. The curve for random rejection is included as a baseline.}
\end{center}
\end{figure}

\begin{figure}[h]
\begin{center}
\vspace*{-5mm}
    \subfloat[spect]{\includegraphics[scale=0.3]{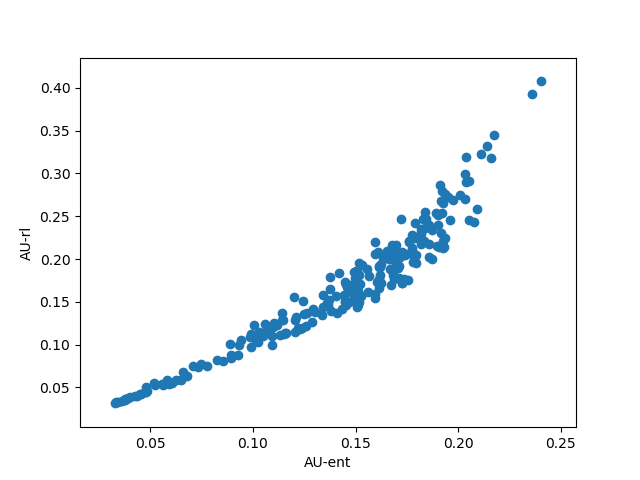} }
  \subfloat[diabetes]{\includegraphics[scale=0.3]{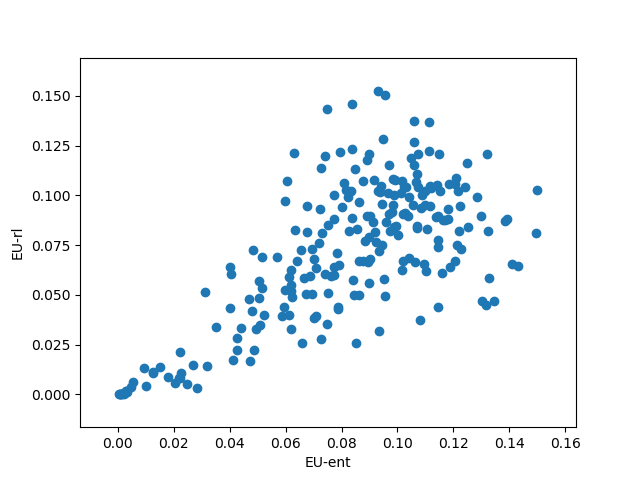}}
\caption{Scatter plot for test set on diabetes data, showing the relationship between the uncertainty degrees (aleatoric left, epistemic right) estimated by the two approaches.}
\end{center}
\end{figure}

As can be seen from Figures 1--4, both approaches to measuring uncertainty are effective in the sense of producing monotone increasing accuracy-rejection curves, and on the data sets we analyzed so far, we could not detect any systematic differences in performance. Besides, rejection seems to work well on the basis of both criteria, aleatoric as well as epistemic uncertainty. This is plausible, since both provide reasonable reasons for a learner to abstain from a prediction. Likewise, there are no big differences between random forests and neural networks, showing that the former are indeed a viable alternative to the latter\,---\,this was actually a major concern of our study. 

\begin{figure}[h]
\begin{center}

\vspace*{-8mm}

  \subfloat[spect]{\includegraphics[scale=0.3]{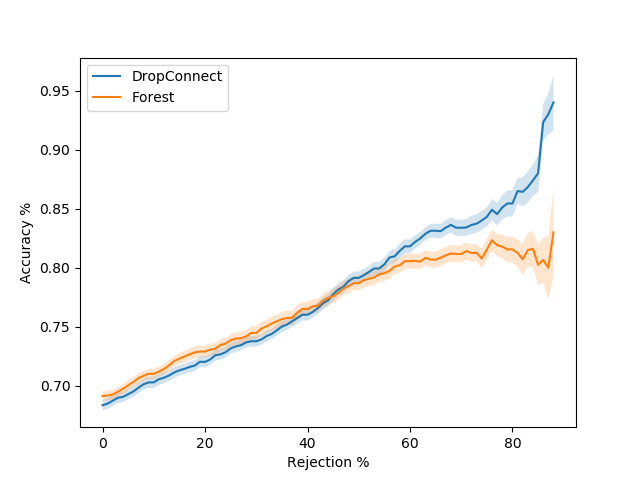} }
  \subfloat[diabetes]{\includegraphics[scale=0.3]{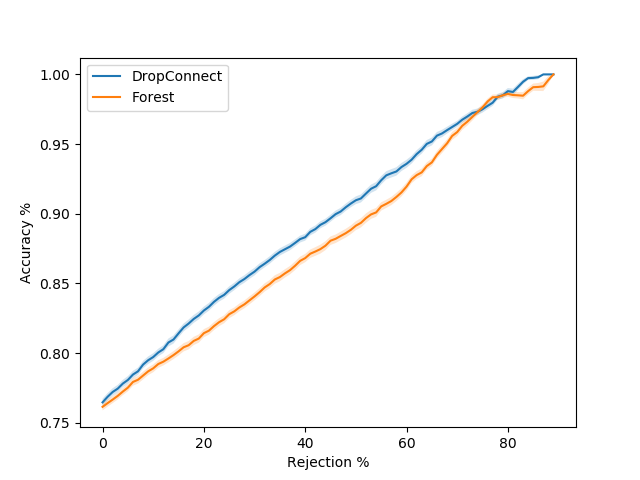}}
  
  \vspace*{-5mm}
  
    \subfloat[spect]{\includegraphics[scale=0.3]{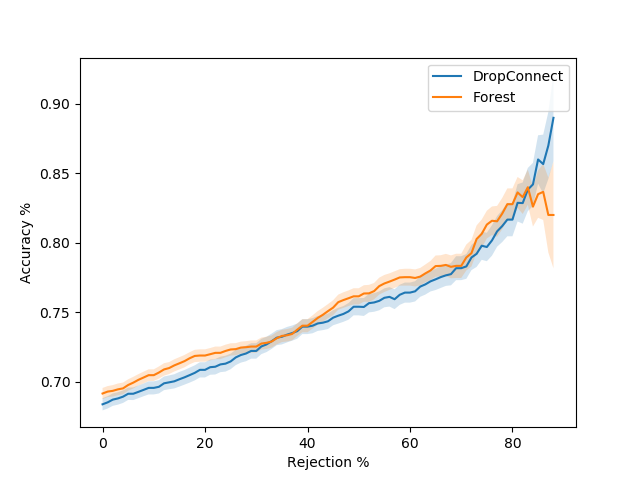} }
  \subfloat[diabetes]{\includegraphics[scale=0.3]{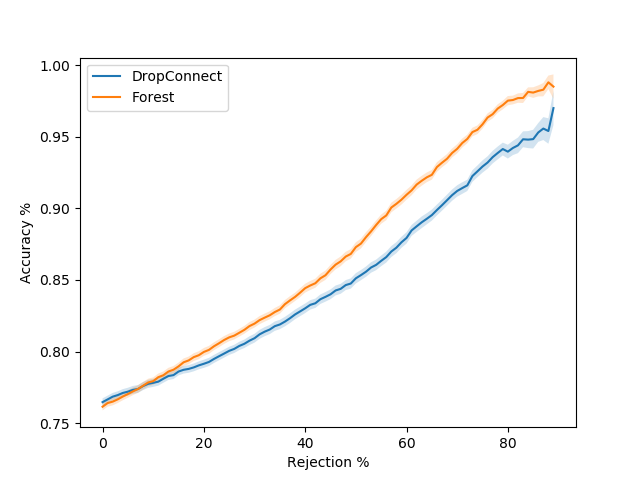}}
  
\caption{Comparison between random forests and neural networks (DropConnect) for aleatoric (above) and epistemic (below) uncertainty.}
\end{center}
\end{figure}

\section{Conclusion}

The distinction between aleatoric and epistemic uncertainty has recently received a lot of attention in machine learning, especially in the deep learning community \cite{kend_wu17}. Roughly speaking, the approaches in deep learning are either based on the idea of equipping networks with a probabilistic component, like in Bayesian deep learning \cite{neal_bl12}, or on using ensemble techniques \cite{laks_sa17}, which can be implemented (indirectly) through techniques such as Dropout \cite{gal_bc16} or DropConnect. The main purpose of this paper was to show that the use of decision trees and random forests is an interesting alternative to neural networks.

Indeed, as we have shown, the basic ideas underlying the estimation of aleatoric and epistemic uncertainty can be realized with random forests in a very natural way. In a sense, they even appear to be simpler and more flexible than neural networks. For example, while the approach based on relative likelihood (Section \ref{sec:rl}) could be realized efficiently for random forests, a neural network implementation is far from obvious (and was therefore not included in the experiments). 

There are various directions for future work. For example, since the hyper-parameters of random forests have an influence on the hypothesis space we are (indirectly) working with, they also influence the estimation of uncertainty degrees. This relationship calls for a thorough investigation. Besides, going beyond a proof of principle with statistics such as accuracy-rejection curves, it would be interesting to make use of uncertainty quantification with random forests in applications such as active learning, as recently proposed in \cite{mpub392}.  


\begin{thebibliography}{10}
\providecommand{\url}[1]{\texttt{#1}}
\providecommand{\urlprefix}{URL }
\providecommand{\doi}[1]{https://doi.org/#1}

\bibitem{birnbaum:1962}
Birnbaum, A.: On the foundations of statistical inference. Journal of the
  American Statistical Association  \textbf{57}(298),  269--306 (1962)

\bibitem{depe_du18}
Depeweg, S., Hernandez-Lobato, J., Doshi-Velez, F., Udluft, S.: Decomposition
  of uncertainty in {B}ayesian deep learning for efficient and risk-sensitive
  learning. In: Proc.\ ICML, 35th Int.\ Conf.\ on Machine Learning.
  Stockholm, Sweden (2018)

\bibitem{gal_bc16}
Gal, Y., Ghahramani, Z.: Bayesian convolutional neural networks with
  {B}ernoulli approximate variational inference. In: Proc.\ of the ICLR
  Workshop Track (2016)

\bibitem{hora_aa96}
Hora, S.: Aleatory and epistemic uncertainty in probability elicitation with an
  example from hazardous waste management. Reliability Engineering and System
  Safety  \textbf{54}(2--3),  217--223 (1996)

\bibitem{mpub170}
H\"uhn, J., H\"ullermeier, E.: {FR3}: A fuzzy rule learner for inducing
  reliable classifiers. IEEE Transactions on Fuzzy Systems  \textbf{17}(1),
  138--149 (2009)

\bibitem{kend_wu17}
Kendall, A., Gal, Y.: What uncertainties do we need in {B}ayesian deep learning
  for computer vision? In: Proc.\ NIPS, pp. 5574--5584 (2017)

\bibitem{krup_pe14}
Kruppa, J., Liu, Y., Biau, G., Kohler, M., K\"onig, I., Malley, J., Ziegler,
  A.: Probability estimation with machine learning methods for dichotomous and
  multi-category outcome: {T}heory. Biometrical Journal  \textbf{56}(4),
  534--563 (2014)

\bibitem{laks_sa17}
Lakshminarayanan, B., Pritzel, A., C.\, Blundell: Simple and scalable
  predictive uncertainty estimation using deep ensembles. In: Proc.\ NeurIPS,
  31st Conference on Neural Information Processing Systems. Long Beach,
  California, USA (2017)

\bibitem{lamb_rc11}
Lambrou, A., Papadopoulos, H., Gammerman, A.: Reliable confidence measures for
  medical diagnosis with evolutionary algorithms. IEEE Trans.\ on Information
  Technology in Biomedicine  \textbf{15}(1),  93--99 (2011)

\bibitem{mobi_dc19}
Mobiny, A., Nguyen, H., Moulik, S., Garg, N., Wu, C.: Drop{C}onnect is
  effective in modeling uncertainty of {B}ayesian networks. CoRR
  \textbf{abs/1906.04569} (2017), \url{http://arxiv.org/abs/1906.04569}

\bibitem{neal_bl12}
Neal, R.: Bayesian learning for neural networks. Springer Science \& Business
  Media  \textbf{118} (2012)

\bibitem{mpub392}
Nguyen, V., Destercke, S., H\"ullermeier, E.: Epistemic uncertainty sampling.
  In: Proc.\ DS 2019, 22nd Int.\ Conf.\ on Discovery Science.
  Split, Croatia (2019)

\bibitem{nguyen:2018}
Nguyen, V.L., Destercke, S., Masson, M.H., H{\"u}llermeier, E.: Reliable
  multi-class classification based on pairwise epistemic and aleatoric
  uncertainty. In: Proc.\ IJCAI, pp. 5089--5095. AAAI Press (2018)

\bibitem{pape_dk18}
Papernot, N., McDaniel, P.: Deep k-nearest neighbors: {T}owards confident,
  interpretable and robust deep learning. CoRR  \textbf{abs/1803.04765v1}
  (2018), \url{http://arxiv.org/abs/1803.04765}

\bibitem{quinlan:1986}
Quinlan, J.R.: Induction of decision trees. Machine learning  \textbf{1}(1),
  81--106 (1986)

\bibitem{safavian:1991}
Safavian, S.R., Landgrebe, D.: A survey of decision tree classifier
  methodology. IEEE transactions on systems, man, and cybernetics
  \textbf{21}(3),  660--674 (1991)

\bibitem{sato_ia18}
Sato, M., Suzuki, J., Shindo, H., Matsumoto, Y.: Interpretable adversarial
  perturbation in input embedding space for text. In: Proceedings {IJCAI} 2018, pp.
  4323--4330. Stockholm, Sweden (2018)

\bibitem{mpub272}
Senge, R., B\"osner, S., Dembczynski, K., Haasenritter, J., Hirsch, O.,
  Donner-Banzhoff, N., H\"ullermeier, E.: Reliable classification: Learning
  classifiers that distinguish aleatoric and epistemic uncertainty. Information
  Sciences  \textbf{255},  16--29 (2014)

\bibitem{vars_es16}
Varshney, K.: Engineering safety in machine learning. In: Proc.\ Inf.\ Theory
  Appl.\ Workshop. La Jolla, CA (2016)

\bibitem{vars_ot16}
Varshney, K., Alemzadeh, H.: On the safety of machine learning:
  {C}yber-physical systems, decision sciences, and data products. CoRR
  \textbf{abs/1610.01256} (2016), \url{http://arxiv.org/abs/1610.01256}

\bibitem{walley:1999}
Walley, P., Moral, S.: Upper probabilities based only on the likelihood
  function. Journal of the Royal Statistical Society: Series B (Statistical
  Methodology)  \textbf{61}(4),  831--847 (1999)

\bibitem{yang_ur09}
Yang, F., Wanga, H.Z., Mi, H., de~Lin, C., Cai, W.W.: Using random forest for
  reliable classification and cost-sensitive learning for medical diagnosis.
  BMC Bioinformatics  \textbf{10} (2009)

\end{thebibliography}

\end{document}